\documentclass[letterpaper]{article} 
\usepackage{aaai25}  
\usepackage{times}  
\usepackage{helvet}  
\usepackage{courier}  
\usepackage[hyphens]{url}  
\usepackage{graphicx} 
\usepackage{natbib}  
\usepackage{caption} 
\usepackage{algorithm}
\usepackage{algorithmic}
\usepackage{paralist}
\usepackage{booktabs}
\usepackage{tabularx}
\usepackage{subcaption}

\usepackage{newfloat}
\usepackage{listings}
\usepackage{censor}
\usepackage{xspace}
\usepackage{amsmath}
\DeclareCaptionStyle{ruled}{labelfont=normalfont,labelsep=colon,strut=off} 
\floatstyle{ruled}
\newfloat{listing}{tb}{lst}{}
\floatname{listing}{Listing}
\def\VexCategory{\textsc{VEXCategory}\xspace}
\def\VexJustification{\textsc{VEXJustification}\xspace}
\def\Internalcomment{\textsc{InternalComment}\xspace}
\def\Customercomment{\textsc{CustomerComment}\xspace}
\def\Vector{\textsc{Vector}\xspace}
\def\Assets{\textsc{Assets}\xspace}
\def\Notifications{\textsc{Notifications}\xspace}
\def\Evaluations{\textsc{Evaluations}\xspace}

\title{CVE-LLM : Ontology-Assisted Automatic Vulnerability Evaluation Using Large Language Models}
\author{
    Rikhiya Ghosh \textsuperscript{\rm 1},
    Hans-Martin von Stockhausen \textsuperscript{\rm 2},
    Martin Schmitt \textsuperscript{\rm 2},\\
    George Marica Vasile \textsuperscript{\rm 3},
    Sanjeev Kumar Karn \textsuperscript{\rm 1},
    Oladimeji Farri \textsuperscript{\rm 1}
}

\affiliations{
    \textsuperscript{\rm 1}Digital Technology and Innovation, Siemens Healthineers USA\\
    \textsuperscript{\rm 2}Cybersecurity, Siemens Healthineers Germany\\
    \textsuperscript{\rm 3}Corporate Technology, Siemens AG Romania\\

    rikhiya.ghosh@siemens-healthineers.com
}

\begin{document}

\maketitle

\begin{abstract}

The National Vulnerability Database (NVD) publishes over a thousand new vulnerabilities monthly, with a projected 25 percent increase in 2024, highlighting the crucial need for rapid vulnerability identification to mitigate cybersecurity attacks and save costs and resources. In this work, we propose using large language models (LLMs) to learn vulnerability evaluation from historical assessments of medical device vulnerabilities in a single manufacturer's portfolio. We highlight the effectiveness and challenges of using LLMs for automatic vulnerability evaluation and introduce a method to enrich historical data with cybersecurity ontologies, enabling the system to understand new vulnerabilities without retraining the LLM. Our LLM system integrates with the in-house application - Cybersecurity Management System (CSMS) - to help Siemens Healthineers (SHS) product cybersecurity experts efficiently assess the vulnerabilities in our products. Also, we present guidelines for efficient integration of LLMs into the cybersecurity tool. 
\end{abstract}

%

\section{Introduction} 
A software vulnerability is a flaw that malicious attackers can exploit to compromise a system's confidentiality, integrity or availability (CIA), such as buffer overflows, SQL injection, and XSS. Regulatory authorities, non-profits, and companies have created open-source standards for sharing vulnerability information, resulting in continually updated databases by cybersecurity professionals. Examples of such databases include: 
\begin{inparaenum}[(a)]
    \item Common Vulnerabilities and Exposures (CVE)\footnote{\url{https://www.mitre.org/}}: Catalogs software vulnerabilities \cite{mitre2005common},  
    \item Common Weakness Enumeration (CWE)\footnote{\url{https://cwe.mitre.org/}}: Describes software security weaknesses, 
    \item Common Attack Pattern Enumeration and Classification (CAPEC)\footnote{\url{https://capec.mitre.org/}}: Details known attack and mitigation patterns, 
    and \item Common Vulnerability Scoring System (CVSS): Assesses vulnerability severity \cite{mell2006common}.
\end{inparaenum}

Among these, the CVE database stands out for its detailed descriptions and assessments of vulnerabilities. The CVE description offers a concise overview of the vulnerability, outlining its nature, affected products, potential impact, and conditions for exploitation. CVSS reports CVE severity in two forms: CVSS Score and CVSS Vector. The CVSS Score, a composite score on a scale of 1 to 10, offers a quick overview of the vulnerability's impact. It is derived from CVSS Vectors, which comprise several metrics evaluating the exploitability of the vulnerability, such as as Base (inherent characteristics), Temporal (changes over time), and Environmental (local CIA requirements and controls). The CVE database includes Base and Temporal metrics, while Environmental metrics are calculated by cybersecurity experts for specific products, ensuring accurate severity assessment within their context.

\begin{table*}
    \begin{tabular}{|p{0.18\linewidth} | p{0.5\linewidth}|p{0.25\linewidth}|}
        \toprule
        \textbf{\Evaluations} &\textbf{Description}&\textbf{Value Type}\\
        \midrule
        \VexCategory & Indicates whether the asset is affected by the notification.& Binary categorical values\\ \hline
        \VexJustification & Provides further explanation if the asset is not affected.&Multiclass Categorical values \\ \hline
        \Internalcomment & Details the problem and advised internal solution.& Text\\ \hline
        \Customercomment & Summarizes the notification's impact and solution directed towards the customer.&Text\\ \hline
        \Vector & Evaluates the CVSS Environmental metrics.&Multiclass, Multilabel categorical values\\
      \bottomrule
    \end{tabular}
    \caption{Details for \Evaluations performed manually currently by SHS cybersecurity experts}
    \label{tab:out_desc}
    \end{table*}

Medical devices are particularly vulnerable to security threats that can compromise patient safety, privacy, medical data integrity, and the availability of diagnostic or treatment devices. According to the US FDA guidance \footnote{Food and Drug Administration (FDA) Guidance on Postmarket Management of Cybersecurity in Medical Devices : \url{https://www.fda.gov/media/95862/download}}, Medical Device Manufacturers (MDMs) must monitor third-party software components for emerging vulnerabilities and assess their impact on the device's safety and security. 

Vulnerability management involves three steps: detection, evaluation, and mitigation. Component vendors detect and issue notifications for vulnerabilities, which MDMs must monitor and assess for relevance to their products. Mitigation may involve vendor patches or MDM-implemented controls. Sometimes, CVE vulnerabilities may not pose a risk, requiring only communication to customers. 

The notifications generated by component vendors consist of one or more CVEs that impact their component. When these notifications are issued, Siemens Healthineers (SHS) Cybersecurity experts conduct manual vulnerability evaluations utilizing CVE descriptions, CVSS scores and the related knowledge in CWE and CAPEC. In SHS, vulnerability evaluation involves the following assessments based on asset details and the notification : 
\begin{inparaenum}[(a)]
    \item \VexCategory, 
    \item \VexJustification,
    \item \Internalcomment,
    \item \Customercomment, and
    \item \Vector.
\end{inparaenum}
Table \ref{tab:out_desc} provides a brief description of these evaluation types.

Given the extensive product portfolios ($\approx$ 1.7K products) of SHS and numerous third-party components, manually performing these evaluations ($\approx$ 1.7 million/month) can be overwhelming. Automation of vulnerability evaluations serves two purposes: 
\begin{inparaenum}[(a)]
    \item quick and efficient management of the large volume of evaluations required, ensuring consistent quality over time, and 
    \item mitigation of the risk associated with the potential unavailability of knowledgeable experts, who may no longer be in the same role or company during the entire post-market lifetime of a device, which can span up to 20 years.
\end{inparaenum}

From a machine learning perspective, SHS vulnerability evaluation tasks in Table \ref{tab:out_desc} could be categorized into two primary tasks: text generation (\Internalcomment and \Customercomment) and text classification (\VexCategory, \VexJustification, and \Vector). Vector generation is a multi-label, multi-class classification problem, \VexCategory is binary classification, and \VexJustification is multi-class classification.

Recent advancements in transformer models\cite{devlin2018bert} have significantly improved language generation and reasoning capabilities across various natural language processing tasks \cite{wu2023autogen, li2023large, yang2024large, sallam2023chatgpt}. While encoder models have traditionally been employed for CVE classification in vulnerability evaluation, the progress in generative AI, particularly decoder-based LLMs, has largely been applied only to penetration testing \cite{happe2023getting, deng2024pentestgpt}. There is a lack of real-world experiments on the effectiveness of LLMs in vulnerability evaluation. To address the challenge MDMs face in assessing a large number of vulnerabilities, we present an LLM trained on historical evaluation data to assess the impact of vulnerabilities on assets. This approach enables faster detection and assessment of exploitable vulnerabilities, leading to quicker mitigation.

In this work, we developed and deployed an LLM to assist product cybersecurity experts in vulnerability evaluation. Overall our contributions are:  
\begin{enumerate}
    \item To the best of our knowledge, this is the first work that explores using LLMs for vulnerability assessment based solely on asset and vulnerability descriptions in medical technology industry setting. 
    \item We benchmarked our model against other fine-tuned open source LLMs and provide insights into best practices for training and deploying a vulnerability LLM. 
    \item Additionally, we demonstrated how incorporating knowledge from vulnerability knowledge graphs into LLM training enhances performance, particularly in deployment settings. 
\end{enumerate}

\begin{figure}[b!]
  \centering
  \centering 
  \fbox{\includegraphics[width=0.98\linewidth]{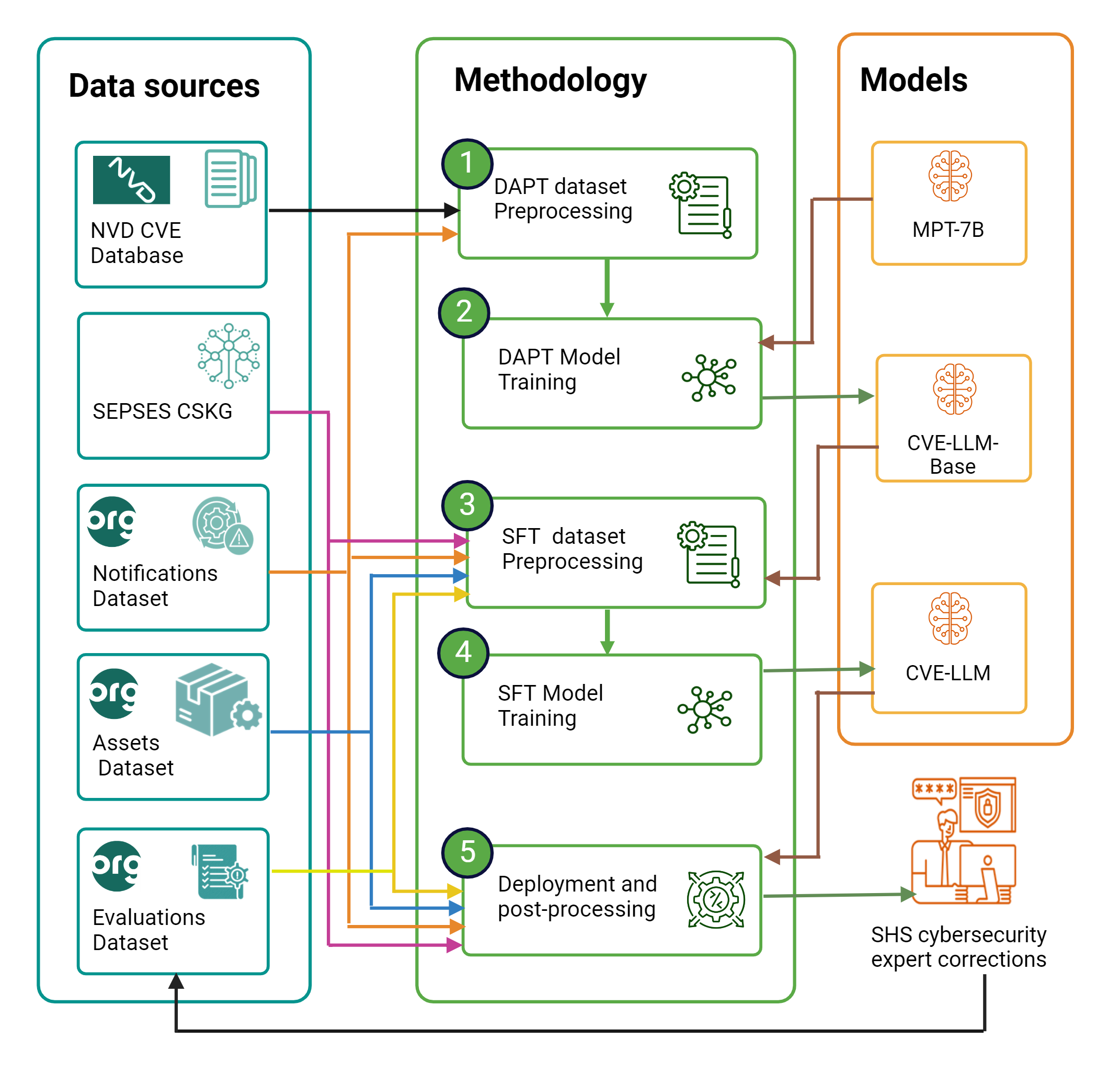}}
      
  \caption{Schematic overview of CVE-LLM}
\end{figure}

\section{Datasets} 
Our methodology adheres to the pretrain-then-finetune paradigm. We conduct Domain Adaptive Pretraining (DAPT) using the open-source MPT-7B model \cite{MosaicML2023Introducing}, which is followed by Supervised Finetuning (SFT) to develop our model, designated as CVE-LLM. Thus, we construct two distinct datasets: the DAPT dataset for the pretraining phase and the SFT dataset for the subsequent fine-tuning phase.

\subsection{DAPT Dataset}
We have formed a DAPT dataset consisting of 350K vulnerability description documents, which combine publicly available CVEs (248K) from the NVD (National Vulnerability Database) and SHS organization-wide vulnerability documents (102K). The organization-wide vulnerability documents describe a combination of one or more vulnerabilities, their effects, and possible mitigations for the affected products. The DAPT dataset includes the CVE title, description, CVSS vector description, affected and unaffected software versions, and mitigation measures (if present). 

The CVE descriptions are cleaned by removing URLs and non-UTF8 characters. We also create a template-based description of the CVSS vectors. For example, for a base vector \emph{AV:L}, the vector description is \emph{Attack Vector is Local}. Details of vectors and their description can be found in FIRST\footnote{\url{https://www.first.org/cvss/v3.1/specification-document}} specification document. 

\subsection{SFT Dataset}

    \begin{table}[!t]
    \begin{center}
    \begin{small}
    \begin{tabular}{m{0.93\linewidth}}
    \hline
    \multicolumn{1}{c}{SFT dataset format}\\
    \hline
    Below is an instruction that describes a task, paired with an input that provides further context.
    Write a response that appropriately completes the request.

    \#\#\# \textbf{Instruction}: \emph{instruction}
    
    \#\#\# \textbf{Input:}
    
    \textbf{Organization:} \emph{sub-organization name}
    
    \textbf{Software:} \emph{software name and version}
    
    \textbf{Product:} \emph{asset name and description}
    
    \textbf{Notification:} \emph{cleaned notification description}
    
    \textbf{Prerequisites:} \emph{cleaned description of Prerequisites from CAPEC}

    \textbf{Typical severity:} \emph{cleaned description of Typical severity from CAPEC}

    \textbf{Mitigations:} \emph{cleaned description of Mitigations from CAPEC}
    
    \textbf{Components present in software:} \emph{description of components common between the asset and the notification}
    
    \textbf{Base and Temporal Vectors:} \emph{description of base and temporal vectors in notification}

    \textbf{CVSS Version:} \emph{CVSS version used in notification}

    \#\#\# \textbf{Response:} \emph{response}
    
    \textless STOP \textgreater
    \\
    \hline
    \end{tabular}
    \end{small}
    \end{center}
    \caption{
    SFT dataset format for a typical instruction}
    \label{tab:instr_format}
    \end{table}

    \begin{table}[!t]
    \begin{center}
    \begin{small}
    \begin{tabular}{m{0.93\linewidth}}
    \hline
\multicolumn{1}{c}{SFT Dataset Example}\\
\hline
    Below is an instruction that describes a task, paired with an input that provides further context.
    Write a response that appropriately completes the request.

    \#\#\# \textbf{Instruction:} Generate internal comments.
    
    \#\#\# \textbf{Input:}
    
    \textbf{Organization:} \censor{DI-DnA}
    
    \textbf{Software:} \censor{Syngo Carbon Monitoring}
    
    \textbf{Product:} \censor{Syngo Carbon Monitoring VB12A}
    
    \textbf{Notification:} Uncontrolled search path in some Intel RST software may allow an authenticated user to potentially enable escalation of privilege via local access. 

    \textbf{Prerequisites:} The attacker must be able to write to redirect search paths on the victim host.

    \textbf{Typical severity:} Very High.

    \textbf{Mitigations:} (1) Implementation: Host integrity monitoring.  (2) Design: Ensure that the program's compound parts, including all system dependencies, classpath, path, and so on, are secured to the same or higher level assurance as the program. (3) Design: Enforce principle of least privilege.
    
    \textbf{Components present in software:} Intel Chipset Device Software - Intel - 10.1.1.44, Intel Graphics Drivers - Intel - 21.20.x, Intel Management Engine Components Installer Driver - Intel - 11.7.0.1043, Intel Network Connections - Intel - 25.0, Intel Trusted Connect Service Client - Intel - 1.47.715.0, Rapid Storage Technology (RST) - Intel - 15.7.x
    
    \textbf{Base and Temporal Vectors:} Attack Vector is Local. Attack Complexity is High. Privileges Required is Low. User Interaction is Required. Scope is Unchanged. Confidentiality is High. Integrity is High. Availability is High. Exploit Code Maturity is Unproven. Remediation Level is Official Fix. Report Confidence is Confirmed.

    \textbf{CVSS Version:} 3.1

    \#\#\# \textbf{Response:} The vulnerability deployed in the system is controlled. Exploitation of this vulnerability requires privileged local access and high Attack Complexity. Exploitability score (0.8) is below threshold. Device access is protected by username and password. The application is executed in Kiosk mode. PII is encrypted in the database. The database cannot be accessed remotely. Firewall rules are configured. The system is protected by whitelisting.
    
    \textless STOP \textgreater
    \\
    \hline
    \hline
    \end{tabular}
    \end{small}
    \end{center}
    \caption{
    SFT dataset example for generating \Internalcomment. The CVE contained in \Notifications for this instruction is CVE-2022-43456. }
    \label{tab:instr_example}
    \end{table}

\begin{table*}
\begin{center}
  \begin{tabular}{ccl}
    \toprule
    Evaluation&Instruction&Response\\
    \midrule
    \VexCategory/\VexJustification & What is the category?&  \emph{Justification}. Category: \emph{Category} \\
    \Internalcomment & Generate internal comment.& \emph{Internal Comment} \\
    \Customercomment & Generate customer comment. & \emph{Customer Comment}\\
    \Vector & Generate environmental vectors. & \emph{Environmental vector description} \\
  \bottomrule
\end{tabular}
\caption{Instructions for each evaluation type}
  \label{tab:instr}
\end{center}
\end{table*}

The SFT dataset is constructed using three organization-wide datasets: 
\begin{inparaenum}[(a)]
    \item \Assets,
    \item \Notifications, and
    \item \Evaluations.
\end{inparaenum}
It is further enriched with knowledge from cybersecurity databases and formatted to perform instruction tuning on the DAPT model.

The \Assets dataset includes all SHS products with details about the software version, third-party components, and associated sub-organizations. The \Notifications dataset is a compilation of vulnerabilities (CVEs) that affect the components, detailing affected components and CVSS base and temporal metrics. The \Evaluations dataset contains expert manual assessments of the impact of a notification on an asset, including \VexCategory, \VexJustification, \Internalcomment, \Customercomment and \Vector.

At the time of writing this paper, there are around 1.7K assets, 145K notifications, and 208K evaluations in \Assets, \Notifications and \Evaluations, respectively. There are 152K unique components used by all the assets, 23K unique CVEs, and 11K unique notifications. Most notification descriptions and internal and customer comments consist of no more than 200, 70, and 50 words, respectively.

\subsubsection{Enriched Instructions Through CSKG}
Cybersecurity Knowledge graphs are known to link disparate but connected cybersecurity resources by means of relationship edges \cite{syed2016uco, iannacone2015developing}. Foremost among these knowledge graphs is the SEPSES which unifies and links vulnerability databases including CVE, CWE and CAPEC into a cybersecurity knowledge graph (CSKG)\cite{kiesling2019sepses}. CSKG is continuously updated and hence the links between all these resources are up to date and current. Utilizing the CSKG, we now enrich \Notifications dataset with prerequisites, mitigations and typical severities of constituent vulnerabilities.

While the CVE entries describe particular instances of vulnerabilities, their corresponding CWE entries describe general types of weaknesses that can lead to these vulnerabilities. The CAPEC database, on the other hand, provides actionable insights into how those weaknesses are exploited through specific attack techniques and patterns. In the CAPEC database, prerequisites refer to the specific conditions or requirements that must be met for an attack pattern to be successfully executed. Typical severity refers to the level of impact or potential damage that an attack pattern can typically inflict if successfully executed. Mitigations refer to strategies, techniques, or practices that can be employed to prevent or reduce the effectiveness of the attack patterns described.

We utilize the SEPSES knowledge graph to identify the CAPEC entries associated with the linked CWE entries relevant to the CVEs included in a notification to enrich the notification description.

\subsubsection{Formation of SFT Dataset}
The format of the SFT dataset, as shown in Table \ref{tab:instr_format}, uses the standard Alpaca format\cite{taori2023alpaca}. Each evaluation in the \Evaluations dataset forms four entries in the SFT dataset. Table \ref{tab:instr} defines the instruction for each evaluation type. An entry in the SFT dataset contains the corresponding instruction, a description of the asset from \Assets dataset, the ontology-enriched description of the notification from \Notifications dataset, the affected software components, CVSS base and temporal metric descriptions, and one of the corresponding assessments: \VexCategory (and \VexJustification), \Internalcomment, \Customercomment, and CVSS \Vector description. CAPEC mitigations are added only for the generation of Internal Comment, while prerequisites and typical severity are added for all evaluation types. The SFT dataset finally goes through text cleaning, removal of incomplete evaluations and long texts (number of tokens \textgreater 1024).

\section{Model Training}

\begin{table*}
\begin{center}
\begin{tabular}{ccccccl}
    \toprule
    Evaluation&Llama3-RAG&Mistral-7B&Llama2-7B&CVE-LLM&CVE-LLM-Eval&CVE-LLM-Prod\\
    \midrule
    \VexCategory & 0.25& 0.64& 0.56&0.94&0.94&0.86\\
    \VexJustification & 0.12&0.60& 0.53&0.90&0.92&0.95\\
    \Internalcomment & 0.19&0.71& 0.62&0.79&0.79&0.73\\
    \Customercomment & 0.28&0.70 & 0.64&0.88&0.89&0.79\\
    \Vector & 0.13&0.59 & 0.57&0.96&0.98&0.98\\
  \bottomrule
\end{tabular}
\caption{Benchmarking CVE-LLM against state-of-the-art Open Source LLMs. CVE-LLM = CVE-LLM results on test dataset (N=44K), CVE-LLM-Eval = deployed CVE-LLM results on test dataset (N=44K), CVE-LLM-Prod = post-deployment results on production(N=10K). Evaluation metrics are Rouge-L score for Internal and Customer comments, and micro-F1 score for the rest: for both metrics, higher numbers indicate better performance.}
\label{tab:llmtest}
\end{center}
\end{table*}

\subsubsection{DAPT Model Training}
The DAPT dataset is randomly split into 90:10 split with 315K training data points and 35K validation data points. We have continuously pretrained MPT-7B base model autoregressively for next token prediction with an objective function of cross entropy. We have expanded the vocabulary of the MPT-7B base model to include names of the components and the organization software, and added 539 new tokens to the 50K vocabulary of MPT. The model is trained using DeepSpeed \cite{rasley2020deepspeed} zero-3 optimization with Lion optimizer \cite{chen2024symbolic} (learning rate: 1e-4, weight decay: 1e-2) for 3 epochs on 8 Tesla A100 SXM4 GPUs.

\subsubsection{SFT Model Training}
 The SFT dataset has 750K unique instructions, and the dataset is randomly split into 600K training, 75K validation, and 44K test data points. The DAPT model is finetuned with the training and validation data by back propagating on completions only, while setting the token labels of the instructions to -100. Except for the learning rate(1e-5), similar optimization, hyperparamaters and GPU resources for training are used for instruction-tuning.

 \section{Deployment}

 The deployment framework includes generation of evaluations using the CVE-LLM, using pre-defined rules to avoid easily identifiable erroneous generations, and using faster inference frameworks to improve model throughput.  The AI-generated vulnerability evaluations are validated by a product cybersecurity expert (with higher priority given to the evaluations with Affected \VexCategory). The corrected human evaluations form the input for subsequent retraining phase.

 \subsubsection{Generation of Evaluations}
 Evaluation generation follows a zero shot model inference where the model is provided a prompt in the pre-defined SFT instruction format, containing information about sub-organization name, software name and version, asset name and version, notification description, CAPEC entries, list of component descriptions for components present in both asset and notification, base and temporal versions and CVSS version. The inference/test dataset is divided into two sub-datasets based on the size of tokens in the inference instruction. For instructions longer than 920 tokens, the sequence length of the trained model is increased to 150 + maximum token length of the longest instruction. If the size of the instruction is less than 920 tokens, the trained model without any changes is used to inference the dataset.

 \subsubsection{Rule-Based Correction}
 Post-processing of model outputs involves using CSMS-specific knowledge in the form of pre-defined rules to correct mistakes in model generation. The rules are as follows: 
 \begin{enumerate}
    \item If \VexCategory value is NotAffected, the evaluation vector is not generated.
    \item If \VexJustification generated is any text other than the pre-defined categories, `Other' gets assigned to \VexJustification.
    \item If \VexJustification is `Other' and Customer Comment is empty, then it is an error in generation. We use Internal Comment as External Comment as well in case it is not empty.
    \item If \VexCategory is Affected, \VexJustification is automatically set to None, and it is ensured that the Vector has the proper CVSS format.

 \end{enumerate}

 \subsubsection{Model Result Integration Into CSMS System}
 The model results are generated daily using vLLM architecture \cite{kwon2023efficient} on one Tesla A100 SXM4 GPU via batch processing, using the newly arrived notifications from third party component vendors and software versions of SHS products. The AI evaluations are generated for every applicable product that uses the third party components relevant to the notifications, as well as for every past notification that is applicable to the product. These evaluations are stored in a database and made available to product cybersecurity experts who can auto-populate the CSMS with AI-recommended evaluations and/or revise the evaluations as needed. The corrected evaluations are subsequently archived for further processing for further training, in addition to investigating mitigation and determining the appropriate reply to customer.

\section{Results}

 The experiment results are reported on two datasets:
 \begin{inparaenum}
     \item Test dataset with 44K evaluations.
     \item Post-deployment dataset results (CVE-LLM-Prod)  with 10K expert evaluations.
 \end{inparaenum}

We use ROUGE-L and micro-F1 for evaluating the model responses: ROUGE-L\cite{lin2004looking} is used for evaluating the model generated \Internalcomment and \Customercomment, whereas micro-F1\cite{scikit-learn} is used for \VexCategory, \VexJustification and (environmental) Vector. 

\subsection{Benchmarking With Open Source LLMs}
We have used open source models that are of similar size as CVE-LLM and are top-performing models on LLM Leaderboards \cite{llm_leaderboard}. We trained them using the DAPT and SFT methods described in the Model Training section. In addition, we also performed RAG over our test dataset with Llama3-70B. Table \ref{tab:llmtest} enumerates the benchmarking results on our test dataset.

CVE-LLM, which is based on MPT-7B model, has shown better performance than Llama2-7B and Mistral-7B-based models. The highest performance improvement is observed for classification-based generations: \VexCategory, \VexJustification and Vector. Post processing the model results have shown the highest improvement for \VexJustification and Vector. We also observe that the performance of CVE-LLM post-deployment (i.e., with completely new assets and notifications) is comparable to the test dataset results. 

\subsection{Ablation Studies}

We conducted ablation studies with respect to different components of model training and inference: 
\begin{inparaenum}[(a)]
    \item DAPT-SFT vs SFT-only system, and
    \item Impact of beam size, temperature, and nucleus sampling during model inference.
\end{inparaenum}
For each of these experiments, the test dataset consists of the same data points from the Evaluations dataset. The training data points have varied across the datasets, but we have made sure that all the possible assets are represented in the training datasets.

\begin{figure*}
    \centering
    \begin{subfigure}[b]{0.31\textwidth}
        \includegraphics[width=\textwidth]{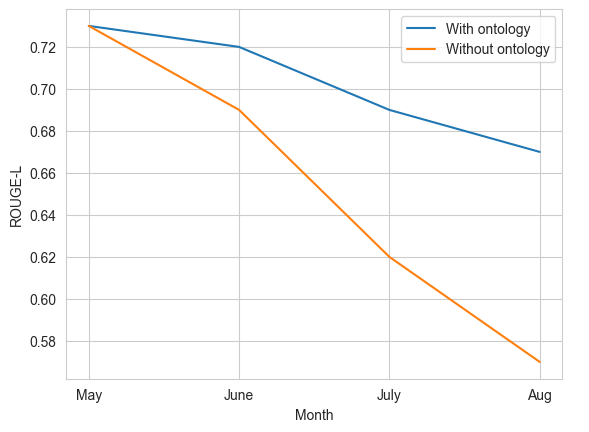}
        \caption{Change in model performance over time in deployment (\Internalcomment)}
        \label{fig:ic}
    \end{subfigure}
    \begin{subfigure}[b]{0.31\textwidth}
        \includegraphics[width=\textwidth]{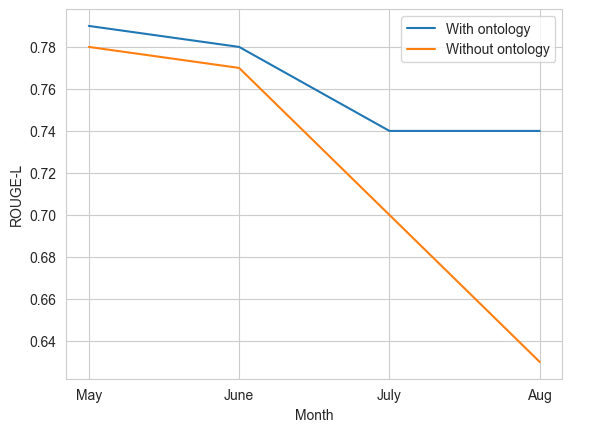}
        \caption{Change in model performance over time in deployment (\Customercomment)}
        \label{fig:cc}
    \end{subfigure}
    \begin{subfigure}[b]{0.31\textwidth}
        \includegraphics[width=\textwidth]{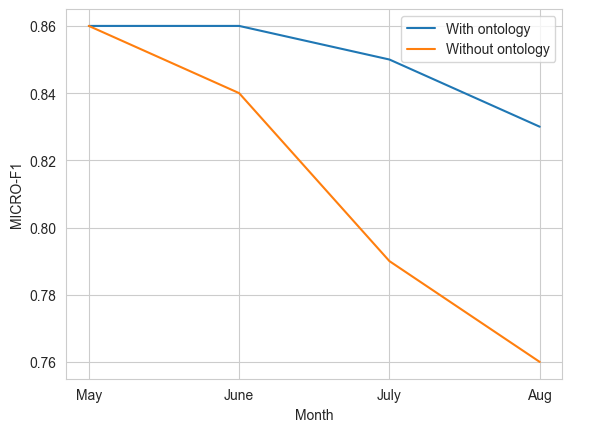}
        \caption{Change in model performance over time in deployment (\VexCategory)}
        \label{fig:vc}
    \end{subfigure}
    \hfill
    \begin{subfigure}[b]{0.31\textwidth}
        \includegraphics[width=\textwidth]{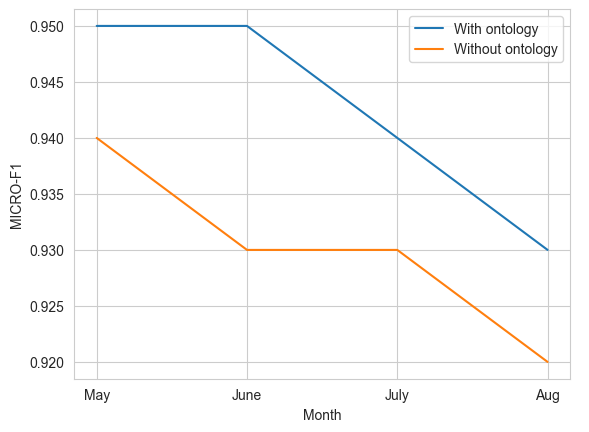}
        \caption{Change in model performance over time in deployment (\VexJustification)}
        \label{fig:vj}
    \end{subfigure}
    \begin{subfigure}[b]{0.31\textwidth}
        \includegraphics[width=\textwidth]{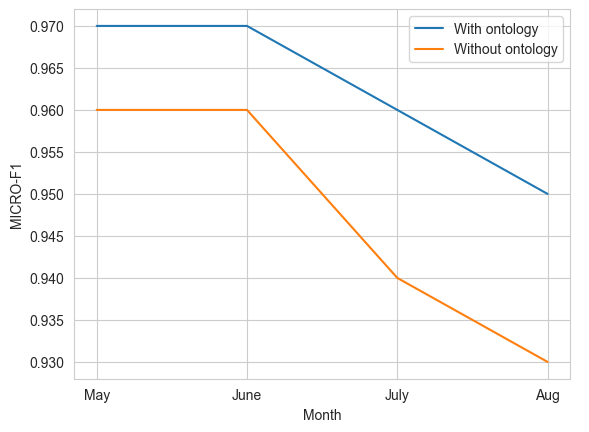}
        \caption{Change in model performance over time in deployment (\Vector)}
        \label{fig:vec}
    \end{subfigure}
    \begin{subfigure}[b]{0.31\textwidth}
        \includegraphics[width=\textwidth]{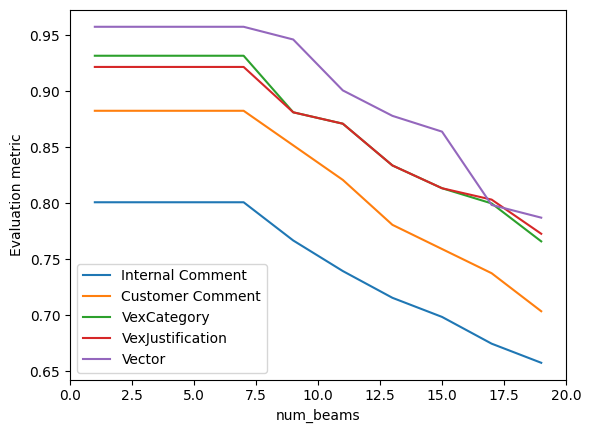}
        \caption{Effect of change of beam size on model performance in test dataset}
        \label{fig:beam}
    \end{subfigure}
    \caption{Ablation studies. The Evaluation metrics is Rouge-L for \Internalcomment and \Customercomment, and micro-F1 for the rest.}\label{fig:ablation}
\end{figure*}

\subsubsection{Effect of Domain Adaptation}

The results after Supervised Finetuning of Domain-adapted CVE-LLM-Base vs Supervised Finetuning of MPT-7B is shown in Table \ref{tab:testdapt}. Domain adaptation and vocabulary expansion leads to a much better performance for all the instruction types, as shown in Table \ref{tab:testdapt}.

\subsubsection{Effect of Inference Parameters}
In generative model inference, parameters such as temperature, beam size, and nucleus sampling significantly influence the quality and diversity of the generated outputs. Temperature controls the randomness of the output, with higher values leading to more diverse results. Nucleus Sampling (top-p sampling) dynamically adjusts the candidate set of tokens to balance between diversity and coherence based on the cumulative probability threshold `p'. Beam Size in beam search influences the depth of search for probable sequences, balancing between computational cost and output quality. Generally these parameters are adjusted to optimize the trade-off  between diversity and coherence of the generated text. We performed inference on our test dataset using CVE-LLM-Eval model to conduct the experiment on the effect of inference parameters.

\textit{Temperature and Nucleus Sampling.} In order to test the effect of temperature, we maintained the generation parameters at their default settings and conducted tests by varying the temperature values from 0 to 1 in increments of 0.1 to evaluate the outcomes. Similarly, for top-p sampling too, we changed the parameter `top-p' from 0 to 1 in increments of 0.1, keeping all other generation parameters in default settings. For both of the experiments, no changes were observed.

\textit{Beam Size.} We measured the effect of beam size for each of the instruction types by changing the beam size from 1 to 19 in increments of 2, maintaining other generation parameters in their default values. We notice that beyond beam size 7, the performance of the system steadily declines for all the instruction types. The effect of change of beam size on model generation is shown in Figure \ref{fig:beam}. 

\subsubsection{Effect of Ontology-Assisted Enrichment.} We have tested the effect of ontology enrichment on our test dataset (Table \ref{tab:testdapt}) as well as deployment data (Figure \ref{fig:ablation}) with 10K evaluations. The model has been trained with data until April and has been deployed since May. The effect in deployment shows the efficacy of the ontology-assisted enrichment.

\begin{table}[b!]
\begin{center}
  \begin{tabular}{cccl}
    \toprule
    Evaluation&CVE-LLM&No&No\\
    &&DAPT&ontology\\
    \midrule
    \VexCategory & 0.94&0.83&0.93\\
    \VexJustification & 0.90& 0.80&0.89\\
    Internal Comment & 0.79&0.64&0.80 \\
    Customer Comment & 0.88 & 0.69&0.88 \\
    Vector & 0.96 & 0.84&0.95\\
  \bottomrule
\end{tabular}
\caption{Test dataset Ablation Results: Results without DAPT and without ontology. The Evaluation metrics is Rouge-L for Internal and Customer comment, and micro-F1 for the rest.}
  \label{tab:testdapt}
\end{center}
\end{table}

\subsection{Inference Time}
    
We have used different techniques to optimize for the time required in inference. The first experiment used popular model serving algorithm vLLM adapted for MPT-7B to find required inference time for one Tesla A100 SXM4 GPU for each of the instruction types and found speedup of 10x magnitude. The experiment results are shown in Table \ref{tab:testinf}. In addition, the adaptive generation by dividing the inference batch into batches of small and large token lengths also improves the average performance over the test dataset to 4.5 second per evaluation from 8 second. The other adaptation of using different maximum sequence length for each instruction type brings down the inference time for each data point in the Evaluations dataset by 1.5 times. We have calculated the time required by a cybersecurity expert in our organization to evaluate vulnerabilities. The mean time required for an expert, averaged over 7K evaluations is 194s with a median of 58s, whereas an evaluation by an our model takes only 1 second per evaluation.

\begin{table}
    \begin{center}
      \begin{tabular}{cccl}
        \toprule
        Evaluation&No speedup&vLLM\\
        \midrule
        \VexCategory & 0.5&0.04\\
        Internal Comment & 2.5&0.20 \\
        Customer Comment & 2& 0.19 \\
        Vector & 3&0.32\\
      \bottomrule
    \end{tabular}
    \caption{Inference time in seconds on our test dataset for model serving with sequence length = 20K.}
      \label{tab:testinf}
    \end{center}
    \end{table}

\section{Discussion}

In this section, we will elucidate our observations related to training and performance of LLMs for vulnerability evaluation, potential areas of improvement and future work. 

\subsubsection{Ontology Enrichment.} While ontology enrichment did not yield substantial performance improvement on the test dataset, it demonstrated a significant impact during deployment on unseen notifications and assets. Most of the errors on unseen notifications pertain to the absence of a linking factor between a new notification and an older similar notification. A close examination show that for most notifications with similar prerequisites, the responses to the instructions are similar across assets. In the future, we would like to experiment with adding more knowledge about the notifications from other security knowledge graphs.

\subsubsection{Domain Adaptation.} We observed significant improvement in performance with domain adaptation of the LLM. Error analysis for the model without domain adaptation shows more hallucinations, more generalized comment generation and lower performance for \VexJustification and Vectors that are less represented. 

\subsubsection{Performance Across Instructions.}  
The performance of CVE-LLM in classification tasks is generally better than the generation tasks. However, though the F1 score for Vector classification is generally high, it remains low for underrepresented classes, resulting in a decline in overall performance. Similarly, performance drops from \VexCategory to \VexJustification classification, largely due to the presence of underrepresented classes in \VexJustification. 

We surmise that the performance of CVE-LLM for \Customercomment is better than that in \Internalcomment due to the following factors:
\begin{itemize}
    \item We observed that approximately 85\% of the customer comments follow a highly templated format and are repeated across multiple assets multiple subdivisions for similar notifications, which is not the case for internal comments.
    \item Internal comments tend to mention specific products, components and versions, which is generally not seen in customer comments. CVE-LLM tends to hallucinate on named entities.
\end{itemize}

\subsubsection{Generation Parameters.} We have noticed model output variations with beam size, with decline in performance with higher beam sizes. Increase in beam search size increases the diversity of text, and that leads to less contextually relevant outputs, as seen in our experiments. Low beam size is sufficient for high quality generation because of the specialized nature of the dataset. The difference between probabilities of less probable outcomes and highly probable outcomes, given the context, is low. This leads to no temperature or nucleus sampling-based variations in model-generated outputs.

\subsection{Error Analysis Across LLMs}
We have used MPT-7B to train CVE-LLM for two reasons:
\begin{inparaenum}[(a)]
    \item Expansion capability to at least 20K tokens without significant loss of performance, and
    \item It performs better than other open source models we tested with our dataset
\end{inparaenum}
However, among all the LLMs we have trained, majority of the issues we have encountered are remarkably similar, differing primarily in their severity. The errors we have encountered across LLMs can be of four types:
\begin{itemize}
    \item Hallucination in named entities: The generated text for both Internal Comment and Customer Comment has been observed to have omitted or falsely included several critical details, particularly those concerning the affected software versions, the recommended update versions, the names of the software or components, and other similarly named entities that constitute essential information. We plan to introduce custom losses during model training in the future that will counteract these hallucinations.
    \item Spurious text generation: This error is more common for Llama2 model and was observed in MPT-7B before introduction of the dedicated STOP token.
    \item Performance degradation due to long text: In spite of having the ability to extend the sequence length of the trained model beyond the training sequence length, the performance on long sequences still falls behind. Long instructions comprises of multiple vulnerabilities, and constitute only 10\% of our test dataset. In the future, we plan to segment it into multiple vulnerabilities and use more robust semantic understanding of each vulnerability to perform vulnerability chaining. We also tested the system only with zero shot instructions at the inference, and we surmise that we can utilize language understanding capabilities better by using few shot methods to assess vulnerabilities for assets.
    \item Errors due to different patterns of task responses in organization subdivisions: The variation of task responses across subdivisions have been seen to affect the quality of generation though the instruction input context includes the subdivision name. This is an issue especially when a notification has vastly more representation in another subdivision. In the future we would attempt to address this issue with data augmentation and sampling techniques.
\end{itemize} 

\section{Relevant Work} 

Language Models have been used extensively in vulnerability management 
\begin{inparaenum}[(a)]
    \item to determine CVSS metrics from CVE description,
    \item to establish a mapping between vulnerabilities in CVE database with the corresponding attack tactics and techniques in ATT\&CK database \footnote{MITRE ATT\&CK database: \url{https://attack.mitre.org/}},
    \item for vulnerability detection, and
    \item for vulnerability repair.
\end{inparaenum} 
Determination of CVSS metrics using vulnerability description has been treated as a text classification/regression problem. Mapping to CVSS score follows a linear regression using Bag-of-Words model \cite{elbaz2020fighting} or neural network model using Doc2Vec \cite{vasireddy2023cvss} method of extracting features from a document. CVSS-BERT \cite{shahid2021cvss}, on the other hand, trained different BERT models for classification of CVE descriptions to the values of different CVE vectors. Both encoder and decoder based models have been used for mapping vulnerability to ATT\&CK tactics and techniques. CVET \cite{ampel2021linking}, a RoBERTa\cite{liu2019roberta}-based model, classified CVE descriptions to one of the ten tactics in ATT\&CK, whereas SMET \cite{abdeen2023smet} used BERT-based textual similarity to map CVE entries to ATT\&CK techniques. Though ChatGPT-based approaches did not yield state-of-the-art results \cite{liu2023not}, VTT-LLM \cite{zhang2024vtt}, trained using various versions of the decoder Bloom model\cite{le2023bloom} with Chain-of-thought \cite{wei2022chain} instructions, incorporated relations between core concepts in CWE \cite{christey2013common} and CAPEC databases \footnote{MITRE CAPEC database: \url{https://capec.mitre.org/}} for CVE to ATT\&CK mapping and surpassed encoder based models. Other work \cite{yosifova2021predicting} involved CVE Vulnerability type classification using TF-IDF and standard machine learning classifier. 

For vulnerability detection, encoder-based LLMs \cite{ameri2021cybert, yin2020apply} have been instrumental, particularly through innovative approaches like the pretrain-and-finetune paradigm. These approaches have also leveraged novel pretraining strategies, the integration Graph Neural Networks\cite{sewak2023crush} or Long Short-Term Memory (LSTM)\cite{hassanin2024pllm} networks, prompt tuning, program analysis, specialized calculus-based causal reasoning, and knowledge graph-based reasoning. In recent times, the focus has shifted towards decoder-only LLMs\cite{zhou2024large} for vulnerability detection. Our work focuses on using pretrain-and-finetune approach with decoder-based LLMs, coupled with ontology enrichment for vulnerability evaluation generation.

\section{Conclusion}
This study illustrates the capacity of large language models (LLMs) to learn from expert-curated historical vulnerability evaluation data, thereby enabling the automation of vulnerability evaluation generation for MDMs. Our model (CVE-LLM) has proven to be effective in learning from past evaluations and has shown high accuracy in prediction of \VexCategory and CVSS Vectors. The model's inference is also substantially faster, achieving speeds approximately 50-100 times greater than those of a human expert. The system is deployed in a human-in-the-loop manner, assisting product cybersecurity experts in swiftly identifying vulnerabilities that affect the assets and communicating mitigations promptly to the customer.

In future work, we intend to investigate additional knowledge infusion techniques, incorporating a broader range of cybersecurity databases. Addressing the challenge of hallucinations remains a key focus. Furthermore, we aim to develop mechanisms for integrating more comprehensive product knowledge by leveraging source code and official cybersecurity documentation. Beyond standardized databases, numerous blogs and websites provide continuous updates on vulnerabilities, which we intend to incorporate into our DAPT and SFT training paradigms. The success of CVE-LLM in the SHS vulnerability evaluation platform has also highlighted opportunities to assist cybersecurity experts in vulnerability mitigation.

\section{Disclaimer}
The concepts and information presented in this paper/presentation are based on research results that are not commercially available. Future commercial availability cannot be guaranteed.

\bibliography{aaai25, main}

\end{document}